\RequirePackage{fix-cm}
\documentclass[twocolumn]{svjour3}           
\smartqed  
\usepackage{graphicx}
\usepackage[utf8]{inputenc}
\usepackage{booktabs}
\usepackage{multirow}
\usepackage{amsmath}
\usepackage{amssymb}
\usepackage{picinpar}  

\journalname{K\"unstliche Intelligenz}  
\begin{document}

\title{Multitask and Multilingual Modelling for Lexical Analysis
}


\author{Johannes Bjerva}


\institute{Dep. of Computer Science,
           Uni. of Copenhagen,
           Denmark \\
           \email{bjerva@di.ku.dk}
}

\date{Received: date / Accepted: date}

\maketitle

\begin{abstract}
In Natural Language Processing (NLP), one traditionally considers a single task (e.g.~part-of-speech tagging) for a single language (e.g.~English) at a time.
However, recent work has shown that it can be beneficial to take advantage of relatedness between tasks, as well as between languages.
In this work I examine the concept of relatedness and explore how it can be utilised to build NLP models that require less manually annotated data.
A large selection of NLP tasks is investigated for a substantial language sample comprising 60 languages.
The results show potential for joint multitask and multilingual modelling, and hints at linguistic insights which can be gained from such models.
\end{abstract}

\section{Introduction}\label{sec:intro}
When learning a new skill, you take advantage of pre-existing skills and knowledge.
For instance, a skilled violinist will likely have an easier time learning to play the cello.
Similarly, when learning a new language you take advantage of the languages you already speak.
For instance, a Norwegian speaker attempting to learn Dutch will likely find their similarities useful.
Recent work has shown that such similarities are also helpful in the context of Natural Language Processing (NLP), which can be defined as the study of computational analysis of human languages.
In this work, I present experiments on multitask and multilingual modelling, i.e., joint learning of several tasks for several languages \cite{bjerva:phd}.

\subsection{Definitions}
\begin{figure*}[t]
  \centering
  \includegraphics[width=0.85\textwidth]{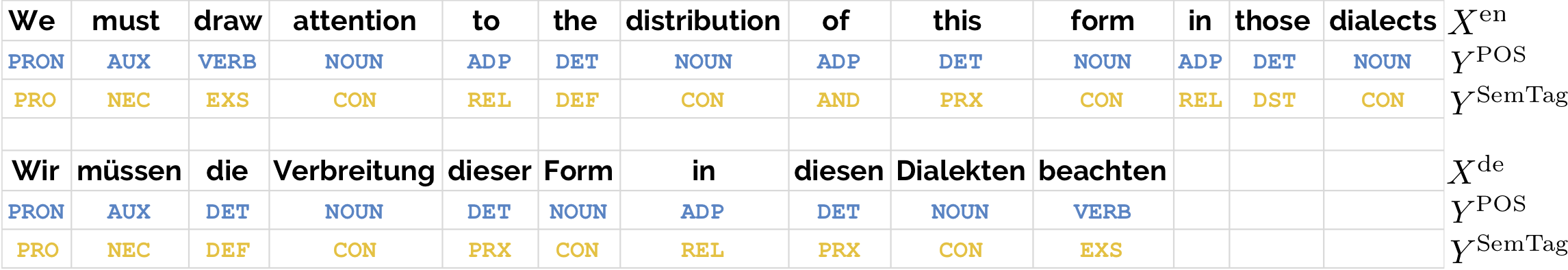}
  \caption{\label{fig:example}Example in English and German with two annotation layers:  semantic tags (SemTag) and parts-of-speech (POS).}
\end{figure*}

\textbf{Lexical Analysis}, the family of tasks under consideration in this work, indicates prediction of linguistically motivated labels for each word in a sentence.
Such labels exist for a number of linguistic levels (e.g.~semantic, syntactic, morphological), and annotated datasets are available for a number of languages.
An example of two layers of annotation for two languages is given in Fig.~\ref{fig:example}.

\textbf{Multitask Learning} (MTL), the main methodological framework in this work, is approached with hard parameter sharing in (deep) neural networks, sharing all parameters except for those in task-specific output layers \cite{mtl}.
Input sentences are defined as sequences $X=(x_0, \cdots, x_n)$, where $x_n\in\mathbb{R}^m$ is a distributed representation of word $n$.\footnote{In NLP words are commonly represented by embedding them in a vector space, typically with $64-256$ dimensions. 
These representations are learnt by predicting contexts in large text corpora, such that words occurring in similar contexts are close to one another, which is useful since such words tend to have similar meanings (i.e.~distributional semantics).}
These sequences are associated with one or more label sequences $Y^t=(y^t_0, \cdots, y^t_n)$, where $y^t_n\in\mathbb{N}$ is the label for $n$ given task $t$ ($Y^{\text{SemTag}}$ and $Y^{\text{POS}}$ in Fig.~\ref{fig:example}).\footnote{SemTags: \cite{pmb,bjerva:2016:semantic}. POS: UD1.3 (universaldependencies.org).}

\textbf{Multilingual Learning} is approached by using multilingual word representations, such that different languages are represented in the same vector space.
Given input sequences in two languages ($X^{\text{en}}$ and $X^{\text{de}}$ in Fig.~\ref{fig:example}), semantically similar words will be highly similar to one another.\footnote{This can be done by learning \textit{multilingual word embeddings}, in which, e.g., the words \textit{dialects} and \textit{Dialekten} are close to one another.}
In addition to the unified input representations, it is typically advantageous to provide the model with a language representation, which encodes the language under consideration.
These representations can be embedded in a separate vector space from the word embeddings, and have been found to encode linguistic features, such as word order \cite{bjerva_augenstein:naacl:18,bjerva_augenstein:17}. 

\subsection{Problem}
Traditionally, NLP practitioners have looked at solving a single task for a single language at a time.
For instance, considerable time and effort might be put into engineering a system for labelling each word in an English sentence with its part of speech (POS), or with a tag representing its semantic content (SemTags).
However, different levels of linguistic analysis tend to exhibit high correlations with one another.
Considering $Y^{\text{SemTag}}$ and $Y^{\text{POS}}$ in Fig.~\ref{fig:example}, the distinctions they make compared to one another in this example are few.
In fact, there are only two apparent systematic differences.
Firstly, SemTags offer a difference between definite (\texttt{DEF}), proximal (\texttt{PRX}), and distal determiners (\texttt{DST}), whereas POS lumps these together as \texttt{DET}.
Secondly, the SemTags also differentiate between relations (\texttt{REL}) and conjunctions (\texttt{AND}), which are both represented by the \texttt{ADP} tag.
This raises the question of how this fact can be exploited, as it is a waste not to take advantage of such inter-task correlations.

Similarly to how different tag sets correlate with each other, languages also share many commonalities with one another.
These resemblances can occur on various levels, with languages sharing, e.g., syntactic, morphological, or lexical features.
A trained linguist comparing the English and German annotations in Fig.~\ref{fig:example} would quickly notice the high correlation between the POS and SemTags used, as well as the high lexical overlap.
As in the case of related NLP tasks, this begs the question of how multilinguality can be exploited.

Finally, given the large amount of data available for many languages in different annotations, it is tempting to investigate possibilities of combining the paradigms of multitask learning and multilingual learning. 
This may allow for, e.g., transferring knowledge across languages and tasks for which limited annotations exist.


\section{Research questions}
The aim of this work is to investigate the following research questions (RQs).
RQ1 and RQ2 deal with MTL, RQ3 and RQ4 with Multilingual Learning, and RQ5 with the combination of the two.
Experiments in this work are run on a large collection of tasks, both semantic and morphosyntactic in nature, and a total of 60 languages are considered, depending on the task.

\newcounter{rtaskno}
\newcommand{\rtask}[1]{\refstepcounter{rtaskno}\label{#1}}
\begin{enumerate}
	\setlength{\itemindent}{.59cm}
    \item [{\bf RQ\ref{rq:stag}}] To what extent can a semantic tagging task be informative for other NLP tasks? \rtask{rq:stag}
    \item [{\bf RQ\ref{rq:mtl}}] How can multitask learning effectivity in NLP be quantified? \rtask{rq:mtl}
    \item [{\bf RQ\ref{rq:sts}}] To what extent can multilingual word representations be used to enable zero-shot learning in semantic textual similarity? \rtask{rq:sts}
    \item [{\bf RQ\ref{rq:multiling}}] In which way can language similarities be quantified, and what correlations can we find between multilingual model performance and language similarities?  \rtask{rq:multiling}
    \item [{\bf RQ\ref{rq:mmmt}}] Can a multitask and multilingual approach be combined to generalise across languages and tasks simultaneously?  \rtask{rq:mmmt}
\end{enumerate}

\subsection*{Semantic Tagging as an Auxiliary Task}

We first look at \textbf{RQ1}, investigating whether a semantic task can be informative for other NLP tasks.
The semantic tag set under consideration consists of 72 tags, and is developed for multilingual semantic parsing \cite{gmb:hla,bjerva:2016:semantic,pmb,bjerva:phd}.
In this study, we use semantic tags as an auxiliary task for the main task, POS tagging.

We use a bi-directional recurrent neural network (bi-RNN), using gated recurrent units.\footnote{Bi-directional RNNs are frequently used in NLP. One advantage of this is that one can use both the preceding and succeeding contexts of a word when predicting its tag.}
The input of the network is a sequence $X=(x_0, \cdots, x_n)$.
Each word, $n$, is represented by a pre-trained word representation, in addition to a word-level character-based representation.
The character-based representation is obtained by running a convolutional neural network (\textsc{ResNet}) over a matrix of character embeddings \cite{bjerva:2016:dsl}.
These representations are concatenated prior to passing them through the bi-RNN.
This allows the model to take advantage of the character-level structure of words, which is beneficial, e.g., in cases where there are unseen words.
For further implementational details, see \cite{bjerva:2016:semantic,bjerva:phd}.

Table~\ref{tab:stag_results} shows that semantic tagging can significantly increase accuracy for POS tagging, thus answering \textbf{RQ1}.
Furthermore, using character representations obtained with a \textsc{ResNet} yields significant improvements above other approaches.

\begin{table}[t]
    \small
\centering
\caption{\label{tab:stag_results}
Results on semtag (ST) and Universal Dependencies (UD) test sets (\% accuracy).
{\sc TNT} indicates a trigram tagger, 
{\sc Bi-lstm} indicates a strong neural baseline, 
{\sc Bi-gru} indicates the $w$ only baseline, 
$w$ indicates usage of word representations,
$c$ indicates usage of char representations, and
the $+${\sc aux} column indicates the usage of an auxiliary task.}
\resizebox{\columnwidth}{!}{
\begin{tabular}{llrrr}
\toprule
& & ST Silver &  ST Gold  & UD v1.3  \\
\midrule
 \multirow{3}{*}{\textsc{Baselines}} 
& {\sc TNT} & 92.09	  &  80.73	   & 92.69  \\
& {\sc Bi-lstm} & 94.98	  &  82.96	   & 95.04  \\
& {\sc Bi-gru} & 94.26	  &  80.26	  & 94.32  \\
\midrule
 \multirow{3}{*}{\textsc{ResNet}} & $c$ & 94.39	  &  76.89	   & 92.63  \\
& $c\land w$ & {\bf95.14}	  &  {\bf 83.64}	     & 94.88  \\
& $+${\sc aux} & 94.23	  &  74.84	  & {\bf 95.67}  \\
 \bottomrule
\end{tabular}
}
\end{table}

\subsection*{Information-theoretic Perspectives on MTL}
In \textbf{RQ2}, we take an information-theoretic perspective to explaining the results from RQ1 \cite{bjerva:2017:mtl}.
Part of the motivation behind this approach is that the entropy of the labels in an annotated text has been hypothesised to be related to the usefulness of that set of labels as an auxiliary task.
We argue that this explanation is not entirely sufficient.
Take, for instance, two tag sets $Y$ and $Y'$, applied to the same sentence and containing the same tags.
Consider now the case where the labels in every sentence using $Y'$ have been randomly reordered.
The entropy of $Y$ and $Y'$ will be the same, but it is unlikely for $Y'$ to be a useful auxiliary task. 

We posit that correlations between tasks ought to be highly indicative of how useful a task is as an auxiliary task.
The mutual information (MI) of two tag sets is a measure of the amount of information that is obtained of one tag set given the other, and can be considered a measure of `correlation' between tag sets.
The MI is
\begin{equation}
      I(Y';Y)=\sum _{y'\in Y'}\sum _{y\in Y}p(y',y)\log\frac{p(y',y)}{p(y')\,p(y)},
\end{equation}
where $y'$ and $y$ are all variables in the given distributions, $p(y',y)$ is the joint probability of variable $y'$ co-occurring with variable $y$, and $p(y)$ is the probability of variable $y$ occurring at all.
MI describes how much information is shared between $Y'$ and $Y$.
Should two tag sets be completely independent from each other, then knowing $Y$ would not give any information about $Y'$.

\begin{table}[t]
    \centering
    \footnotesize
    \caption{\label{tab:mtl_results}Correlation scores and associated $p$-values, between change in accuracy ($\Delta_{acc}$) and entropy ($H(Y)$), and mutual information ($I(X;Y)$), calculated with Spearman's $\rho$.}
    \begin{tabular}{lrr}
        \toprule
        {\bf Condition} & $\rho(\Delta_{acc}, H(Y))$ & $\rho(\Delta_{acc}, I(X;Y))$ \\
        \midrule
        Full overlap     & $-$0.06 (p$=$0.214)      & 0.08 (p$=$0.114) \\
        Partial overlap  & 0.07  (p$=$0.127)        & \bf{0.43 (p$\ll$0.001)} \\
        No overlap       & 0.08  (p$=$0.101)        & \bf{0.41 (p$\ll$0.001)} \\
        \bottomrule
    \end{tabular}
\end{table}

Experiments are run on 39 of the languages found in UD1.3, with three levels of overlap between the main and auxiliary datasets: full overlap, partial overlap, and no overlap.
We consider POS tagging as a main task, and the more difficult syntactic task of dependency relation classification as an auxiliary task \cite{bjerva:2017:mtl}.
Table~\ref{tab:mtl_results} shows that mutual information is a much better predictor of gains when using an auxiliary task, than entropy.
This is the case when there is some or no overlap between datasets, but not when the datasets are identical, in which case similarity between tasks is no longer a useful predictor.
Hence, when using an auxiliary task dataset which is (partially) separate from the main task dataset, MI can be used to quantify MTL effectivity.

\subsection*{Multilingual Learning in NLP}

We now turn to the research questions dealing with multilingual modelling.
In order to answer \textbf{RQ3}, we train a system on cross-lingual semantic textual similarity \cite{bjerva:14:semeval}.
In short, given two input sentences $X^{l}_n$ and $X^{l}_m$, the objective is to quantify how similar these are to one another on a continuous scale.
In order to investigate zero-shot learning\footnote{Evaluation of a model trained on one language on a test instance for an unobserved language.} in this context, we use multilingual word embeddings, training on English sentence pairs $X^{\text{en}}_n$ and $X^{\text{en}}_m$, and evaluating on Spanish sentence pairs $X^{\text{es}}_n$ and $X^{\text{es}}_m$.
Results indicate that this approach is feasible, but does not yield competitive results \cite{bjerva:2017:sts}.

\begin{figure}[t]
	\centering
	\includegraphics[width=0.85\columnwidth]{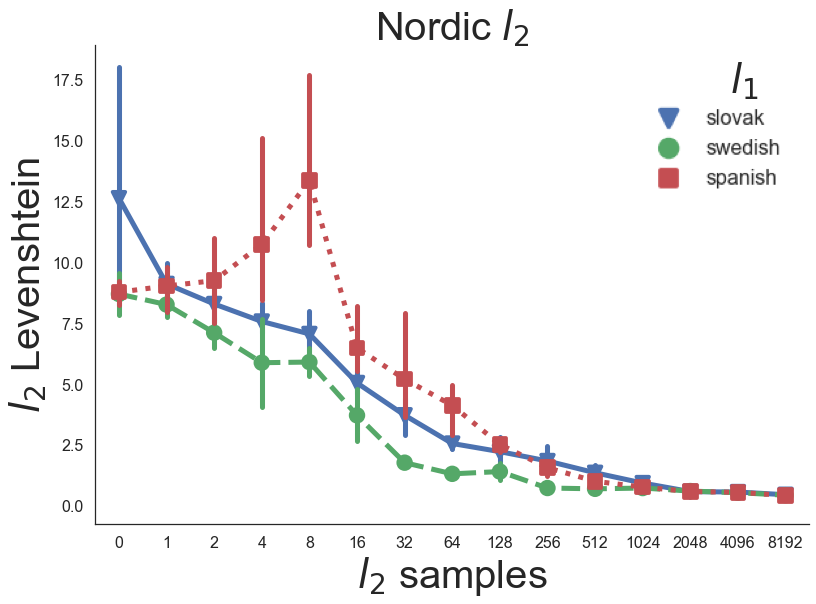}
    \caption{\label{fig:multilingual}Multilingual modelling, training on three source languages ($l_1$) and evaluating on Nordic target languages ($l_2$). The x-axis indicates n $l_2$ samples used. The y-axis indicates system performance on the $l_2$ (lower is better).}
\end{figure}

We now investigate \textbf{RQ4}, looking at how similar languages need to be in order for multilingual modelling to be feasible.
The main finding here is that this type of modelling, using multilingual word embeddings and hard parameter sharing, is only feasible in cases where the languages under consideration are highly similar to one another.
Fig.~\ref{fig:multilingual} shows an attempt at this type of modelling, indicating that transferring from a closely related language (Swedish) is more beneficial than using more distantly related ones (Spanish and Slovak).

\subsection*{Joint Multitask and Multilingual Learning}
Finally, we turn to joint multitask and multilingual modelling, to answer \textbf{RQ5}.
Experiments are run by training on task--language pairs, e.g., $X^{\text{en,fr}}\rightarrow Y^{\text{POS}}$ and $X^{\text{en}}\rightarrow Y^{\text{SemTag}}$ (i.e.~holding out French), and evaluating on the held out $X^{\text{fr}}\rightarrow Y^{\text{SemTag}}$. 
These experiments show some gains for similar languages, and highlight the limits of hard parameter sharing.

The experiments presented here show the benefits of exploiting similarities between tasks and languages, both separately and jointly.
Future work should explore sharing paradigms in which similarities between languages and tasks are exploited in a more structured way, such that heavy sharing is utilised between similar languages as in \cite{delhoneux:2018}, while limiting sharing between less similar ones.


\bibliographystyle{spmpsci}      
\bibliography{thesis}   

\end{document}